# Uniform Solution Sampling Using a Constraint Solver As an Oracle[*]

**Stefano Ermon**
Department of Computer Science
Cornell University
ermonste@cs.cornell.edu

**Carla Gomes**
Department of Computer Science
Cornell University
gomes@cs.cornell.edu

**Bart Selman**
Department of Computer Science
Cornell University
selman@cs.cornell.edu

## Abstract

We consider the problem of sampling from solutions defined by a set of hard constraints on a combinatorial space. We propose a new sampling technique that, while enforcing a uniform exploration of the search space, leverages the reasoning power of a systematic constraint solver in a black-box scheme. We present a series of challenging domains, such as energy barriers and highly asymmetric spaces, that reveal the difficulties introduced by hard constraints. We demonstrate that standard approaches such as Simulated Annealing and Gibbs Sampling are greatly affected, while our new technique can overcome many of these difficulties. Finally, we show that our sampling scheme naturally defines a new approximate model counting technique, which we empirically show to be very accurate on a range of benchmark problems.

## 1 Introduction

In recent years, we have seen significant interest in probabilistic reasoning approaches that combine both hard, logical constraints and probabilistic information through soft, weighted constraints. Markov Logic Networks [1] are a prominent example of such a modeling language. The hard constraints are used to capture definitional and other irrefutable relationships in the underlying domain, while soft constraints capture less categorical information, generally better for modeling real-world data and dependencies.

Markov Chain Monte Carlo (MCMC) methods, such as Simulated Annealing (SA) and Gibbs Sampling [2], are among the most prominent approaches to probabilistic reasoning, especially when exact inference is beyond reach. In fact, they are guaranteed to asymptotically converge to a stationary distribution that can theoretically provide uniform samples. However, sampling from the solutions of a set of hard constraints is believed to be very hard in worst case, as it is closely related to #-P complete problems such as model counting [3, 4, 5, 6, 7]. In particular, the time required for a Markov Chain to reach the stationary distribution (mixing time) is often exponential in the number of problem variables when hard constraints are present. These difficulties led to the introduction of alternate sampling strategies. For example, in the SampleSAT approach [8], a Markov Chain is constructed that combines moves proposed according to a Simulated Annealing Markov Chain with so-called WalkSAT moves, inspired by local search constraint solvers. The SampleSAT method provides a significant advance, since SA sampling on constraint problems of practical interest generally does not reach any solutions or, at best, only a small subset of all possible solutions. SampleSAT was subsequently incorporated into MC-SAT in the Alchemy package for reasoning and learning for Markov Logic Networks [9]. A key limitation of SampleSAT is that it is not guaranteed to converge to the uniform distribution on the solution set. In fact, one can create examples where the stationary distribution of SampleSAT is arbitrarily biased (non-uniform) in terms of solution samples. Still, this does not negate the value of SampleSAT in many practical settings. For, although SA-like sampling does sample in the limit from the uniform stationary distribution on the solution set, reaching the stationary distribution often requires exponential time, making the SA strategy of little use in practice.

In this paper, we revisit the question of how to devise practical methods for sampling from solutions defined by a set of hard constraints on a combinatorial space. We present a series of challenging domains that reveal the difficulties introduced by such constraints. These include high energy[1] barriers, large energy plateaus ("golf-course" like energy landscapes), and highly asymmetric sampling spaces. We present data demonstrating that SA, Gibbs sampling, and SampleSAT [8] are all greatly affected by these problems.

---

[*]This work was supported by NSF Grant 0832782.

[1]Energy is defined as the number of violated constraints.

However, we also show that highly structured energy landscapes can actually present new opportunities for solution samplers. Combinatorially defined energy functions have a rich structure embedded, but traditional MCMC methods are very general and therefore treat the energy as a "black-box", effectively ignoring the underlying structure. On the other hand, modern day constraint solvers use clever heuristic to exploit constraint structure as much as possible, and can solve very large structured industrial problems with millions of variables [10]. However, these solvers cannot be used directly as solution samplers because they tend to oversample certain solutions, as they are designed just to find one satisfying assignment but not to be uniform [8].

In this paper, we propose a novel sampling scheme called SearchTreeSampler, which leverages the reasoning power of a systematic constraint solver while enforcing a uniform exploration of the search space. Constraint solvers have been previously applied by SampleSearch [11, 12, 13] in the context of importance sampling, a framework where the performance is known to heavily depend on the choice of the proposal distribution. In contrast, SearchTreeSampler introduces a new way of exploring the search space that does not rely on a heuristically chosen proposal distribution, and directly provides (approximately) uniform samples. The constraint solver is used as a black-box, so that any systematic solver can be plugged in, with no modifications required. We empirically demonstrate that by leveraging constraint structure, SearchTreeSampler can overcome many of the difficulties encountered by other solution samplers. In particular, we show it can be orders of magnitude faster than competing methods, while providing more uniform samples at the same time. Further, we show that our sampling scheme naturally defines a new technique for approximately counting the number of distinct solutions (model counting), that we empirically show to be very accurate on a range of benchmark problems.

## 2 Problem Definition

We consider the problem of sampling from a combinatorial search space defined by $n$ Boolean variables and $m$ constraints specified by a Boolean formula $F$ in conjunctive normal form (CNF). A constraint or *clause* $C$ is a logical disjunction of a set of (possibly negated) variables. A formula $F$ is said to be in CNF form if it is a logical conjunction of a set of clauses $\mathcal{C}$.

We define $V$ to be the set of propositional variables in the formula, where $|V| = n$. A variable assignment $\sigma : V \to \{0, 1\}$ is a function that assigns a value in $\{0, 1\}$ to each variable in $V$. As usual, the value 0 is interpreted as FALSE and the value 1 as TRUE. Let $F$ be a formula in CNF over the set $V$ of variables with $m = |\mathcal{C}|$ clauses and let $\sigma$ be a variable or truth assignment. We say that $\sigma$ satisfies a clause $C$ if at least one signed variable of $C$ is TRUE. We say that a truth assignment $\sigma$ is a satisfying assignment for $F$ (also called a model or a solution) if $\sigma$ satisfies all the clauses $C \in \mathcal{C}$. Let $\mathcal{S}_F$ be the set of solutions of $F$, and let $Z = |\mathcal{S}_F|$ be the number of distinct solutions.

Given a Boolean formula $F$, we define a discrete probability distribution $D$ over the set of all possible truth assignments $\{0, 1\}^n$ such that

$$D(\sigma) = \begin{cases} 1/Z & \text{if } \sigma \in \mathcal{S}_F \text{ (i.e., } \sigma \text{ is a solution)} \\ 0 & \text{otherwise} \end{cases}$$

In this paper we consider the problem of sampling from $D$. This problem is very hard and in fact simply deciding whether or not the support of $D$ is empty is NP-complete (the CNF-SAT problem). Sampling is however believed to be even harder, as it is closely related to #-P complete problems such as inference and model counting [3, 4]. For instance, SAT solvers cannot be directly used as solution samplers because they tend to oversample certain solutions, as they are designed just to find one satisfying assignment but not to be uniform [8].

## 3 Background on Solution Sampling

In this section, we briefly describe the main techniques for solution sampling.

### 3.1 Simulated Annealing

Simulated Annealing is a MCMC algorithm that defines a reversible Markov Chain on the space of truth assignments $\{0, 1\}^n$ to sample from a Boltzmann distribution [14]. The transition probabilities (and the steady state probability distribution) depend only on a property of the truth assignments called "energy". The energy $E : \{0, 1\}^n \to \mathbb{N}$ gives the number of clauses violated by a truth assignment $\sigma$ and is defined as follows

$$E(\sigma) = |\{c \in \mathcal{C} | \sigma \text{ does not satisfy } c\}|.$$

The Boltzmann steady state probability distribution is given by

$$P_T(\sigma) = \frac{1}{Z(T)} e^{-\frac{E(\sigma)}{T}},$$

where $T$ is a formal parameter called "temperature", and $Z(T)$ is the normalization constant. Notice that $P_T$ assigns the same probability to all solutions, i.e. $P_T(\sigma) = \alpha$ for all $\sigma \in \mathcal{S}_F$, but is not necessarily zero for non-solutions. Given an algorithm $\mathcal{A}$ that produces samples from $P_T$, we can construct an algorithm $\mathcal{B}$ that samples from $D$ (i.e. uniformly from the solution set) using rejection sampling. More specifically, we take samples $s$ produced by $\mathcal{A}$ and we discard all the ones such that $s \notin \mathcal{S}_F$. The fundamental tradeoff involved is that we want the probability distribution $P_T$ to be easier to sample from (compared to $D$), but at the same time the probability mass should be concentrated on satisfying assignments (i.e. $P_T$ should be close enough to $D$) so that we don't generate too many unwanted samples (i.e. non-solutions). Closely related to Simulated Annealing (SA) is the Gibbs Sampler for the Boltzmann distribution, that defines a similar Markov Chain with the same steady state probability distribution [2]. In our analysis below, we consider a fixed temperature annealing where $T$ is fixed during the run of the Markov Chain.

### 3.2 SampleSAT

In [8] Wei et al. propose the use of an hybrid approach where they interleave SA moves with moves based on a focused random walk procedure (inspired by local search SAT solvers [15]). This approach is based on a result by Papadimitriou [16] proving that a random walk procedure for SAT will find a solution to any satisfiable 2CNF formula in $O(n^2)$ time, where $n$ is the number of variables in the formula. While the role of focused random walk moves is to find solution "clusters", the role of SA moves is to heuristically provide some level of uniformity of the sampling. The drawback of this approach is that it does not maintain any detailed balance equation so there is no control on the resulting steady state probability distribution. SampleSAT therefore loses all the theoretical guarantees on the uniformity of the sampling provided by SA, and as we show below in the experimental section, it can lead to poor uniformity also in practice.

### 3.3 SampleSearch

SampleSearch [11, 12, 13] is an importance sampling technique that draws samples from the so-called *backtrack-free distribution*. Although it is not a uniform sampler, these samples can be used to compute expectations (using importance sampling) by correcting for the non uniformity of the *backtrack-free distribution*. The support of the *backtrack-free distribution* corresponds exactly to the set of solutions, and this greatly reduces the number of rejected samples. However, as other importance sampling schemes, the performance is highly dependent on the choice of the proposal distribution, which is usually precomputed using a generalized belief propagation scheme [11]. Sampling from the *backtrack-free distribution* can be achieved either by using a complete solver as a black-box, or by using the SampleSearch scheme, which integrates backtracking search with sampling [12]. Our approach is similar in the sense that we also use a complete solver as an oracle as we explore the search tree. However, the search tree is explored is a very different way. Specifically, our *recursive* approach aims at uniformly exploring of the search tree level by level, and directly provides (approximately) uniform samples without need for a heuristically chosen proposal distribution (e.g., using variational methods). We will compare the performance of the two methods for model counting below.

## 4 Black-Box Sampling

Modern day SAT solvers are very effective at finding *a solution*, but they explore the search space in a highly non-uniform way. This is because they use heuristics such assigning "Pure literals" (e.g., if a variable only appears with positive sign in a formula, then it can be safely set to true) that heavily bias the search towards certain parts of the search space.

Modifying a pure DPLL-style algorithm to produce uniform samples is a challenging task. Suppose the choice of the ordering in which variables are set is chosen uniformly at random, and the polarity (whether to assign true or false to a variable during search) is also chosen uniformly at random. Then in general the first solution found is not a uniform sample from $S_F$. A simple counterexample is the formula $x_1 \vee x_2$, where it can be seen that the solution $x_1 = x_2 = 1$ is less likely to be found in both possible variables orderings. In order to obtain uniform samples, one could choose the polarity of the variables according to their marginal probabilities (with respect to $D$), but this would defy the purpose of sampling, since we often want to obtain samples precisely to estimate quantities such as marginals.

Instead of modifying an existing search procedure to produce uniform samples, we introduce a novel recursive sampling scheme that aims at enforcing a level by level uniform exploration of the search tree, while leveraging the reasoning power of a complete SAT solver.

### 4.1 A Recursive Sampling Strategy

Our method is based on the notion of *pseudosolution*, defined as follows:

**Definition 1.** *Let $\pi$ be an ordering of the variables. A* pseudosolution *of level $i$ is a truth assignment to the first $i$ variables that can be completed to form a solution (i.e. a node in the search tree at level $i$ that has a solution as a descendant). We denote $S_i$ the set of* pseudosolutions *of level $i$.*

Our recursive strategy (Algorithm 1) is based on the idea of dividing the search tree into $L$ levels, and recursively sampling *pseudosolutions* using previously generated samples of *pseudosolutions* of a higher level (see Figure 1). In other words, we assume to have access to uniform samples of ancestors of solutions at level $i$, and we generate samples of ancestors of solutions at level $i + \ell$ using Algorithm 2. The procedure is initialized with a pseudosolution of level 0, i.e. an empty variable assignment. Note that generating samples from *pseudosolutions* of level $n$ (if there are $n$ variables) is equivalent to the original problem of sampling solutions from $S_F$.

A complete SAT solver is used in Algorithm 2 to generate

**Algorithm 1** SearchTreeSampler($F, k, M$)

    **Input:** Formula $F$ with $n$ vars. Parameters $k$, $M = 2^\ell$
    **Output:** A set $S$ of solutions of $F$
    **if** $F$ is not satisfiable **then**
        return $\emptyset$
    **else**
        Let $\Phi_0 = \{\top\}$ // True, empty variable assignment
        Let $L = \lceil \frac{n}{\ell} \rceil$ // number of levels
        **for** $i = 1, \cdots, L$ **do**
            $\Phi_i \leftarrow$ BlackBoxSampler($\Phi_{i-1}, k, \ell$)
        **end for**
        return $\Phi_L$
    **end if**

**Algorithm 2** BlackBoxSampler($\Phi, k, \ell$)

    **Input**: A set $\Phi$ of uniformly sampled pseudosolutions of level $i$. Parameters $k, \ell$
    **Output**: A set $S$ of pseudosolutions of level $i + \ell$ (approximately uniformly sampled) with $2^\ell|\Phi| \geq |S| \geq |\Phi|$
    $S = \emptyset$
    **for** $j = 1, \cdots, \min\{k, |\Phi|\}$ **do**
        Sample $s_j$ from $\Phi$ without replacement
        Generate $D(s_j)$, the set of all pseudosolutions of level $i+\ell$ with $s_j$ as ancestor (using a complete SAT solver)
        $S = S \cup D(s_j)$
    **end for**
    return $S$

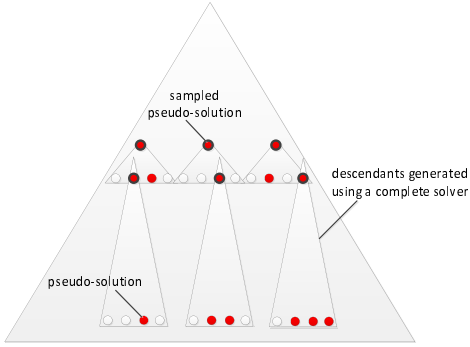

Figure 1: Representation of how Algorithm 1 explores the search tree.

$D(s_j)$, the set of all pseudosolutions of level $i + \ell$ with $s_j$ as ancestor. This is accomplished by repeatedly checking the satisfiability of $F \wedge s_j \bigwedge_{d \in D} \neg d$ until it is provably unsatisfiable, adding a *pseudosolution* $d$ to $D$ each time one is found. Completeness is required in order to enumerate all elements of $D(s_j)$, i.e. to prove unsatisfiability when all elements have been found.

### 4.2 Analysis

It is easy to verify by induction that the sets $\Phi_i$ in Algorithm 1 satisfy the property $|\Phi_i| \geq \min\{k, |S_i|\}$. The key property of Algorithm 2 is that the set $S$ returned is such that sampling from $S$ is approximately equivalent to sampling from $S_{i+\ell}$, the set of *pseudosolutions* of level $i+\ell$. The parameter $k > 1$ controls the uniformity of the sampling. This is formalized by the following Theorem:

**Theorem 1.** *Let $S$ be the output of Algorithm 2 with input $\Phi$ such that $|\Phi| \geq \min\{k, |S_i|\}$, and $s, s' \in S_{i+\ell}$ be any two pseudosolutions of level $i + \ell$. We have*

$$\frac{k}{M+k-1} \leq \frac{P(s)}{P(s')} \leq \frac{M+k-1}{k}, \quad (1)$$

*where $M = 2^\ell$ and $P(s)$ is the probability that a uniformly sampled element from $S$ is equal to $s$.*

*Proof.* Suppose $k \leq |\Phi|$ (otherwise, it means that $\Phi$ contains all pseudosolutions of level $i$, hence by definition $S = S_{i+\ell}$ and therefore $P(s) = P(s')$), and that $\Phi \subseteq S_i$ is a set of uniformly sampled pseudosolutions of level $i$. We can think of each pseudosolution $s \in \Phi$ as an urn, that contains a certain number $1 \leq |D(s)| \leq M = 2^\ell$ of pseudosolutions at a lower level $i + \ell$ (its descendants). Let $|S_i| = N \geq |\Phi|$ be the total number of pseudosolutions of level $i$.

Since $\Phi$ contains uniform samples, $s_1, \cdots, s_k$ in Algorithm 2 are also uniform samples of *pseudosolutions* of level $i$. Let $S$ as in Algorithm 2 (the union of the contents of the $k$ urns selected). Let $s \in S_{i+\ell}$ be a pseudosolution of level $i + \ell$, and let $a(s) \in S_i$ be its unique ancestor at level $i$. Clearly, the probability that $s \in S$ is $P[s \in S] = \frac{\binom{N-1}{k-1}}{\binom{N}{k}}$ that is equal to the probability of selecting the ancestor $a(s)$ of $s$ on level $i$. Let $e$ be a randomly selected element of $S$. Ideally, we would like the probability $P(s) \triangleq P[e = s]$ to be a constant independent of $s$ (uniform sampling). However, intuitively this is not exactly constant because there is a bias towards elements such that $|D(a(s))|$ is small. Specifically,

$$P(s) = \frac{1}{\binom{N}{k}} \sum_{\ell_1 \cdots \ell_{k-1}} \frac{1}{|D(a(s)) \cup D(s_{\ell_1}) \cdots \cup D(s_{\ell_{k-1}})|}.$$

Since the sets are disjoint,

$$P[e=s] = P(s) = \frac{\binom{N-1}{k-1}}{\binom{N}{k}} \sum_{t=1}^{\binom{N-1}{k-1}} \frac{1}{\binom{N-1}{k-1}} \frac{1}{|D(a(s))| + z_t},$$

where $z_t = |D(s_{\ell_1}) \cup \cdots \cup D(s_{\ell_{k-1}})| \geq k - 1$. Let $s' \in S_{i+\ell}$ be another pseudosolution of level $i + \ell$. We rewrite $P(s)$ as

$$\frac{1}{\binom{N}{k}} \left( \sum_{t=1}^{\binom{N-2}{k-2}} \frac{1}{|D(a(s))| + a_t + |D(a(s'))|} + \sum_{t=1}^{\binom{N-2}{k-1}} \frac{1}{|D(a(s))| + b_t} \right).$$

Suppose wlog that $|D(a(s))| \geq |D(a(s'))|$. Then,

$$\frac{1}{|D(a(s'))|+b_t} \leq \frac{|D(a(s))|+k-1}{|D(a(s'))|+k-1}\frac{1}{|D(a(s))|+b_t}$$

by Lemma 1, and using the fact that $1 \leq |D(a(s))| \leq M$,

$$\frac{1}{|D(a(s'))|+b_t} \leq \frac{M+k-1}{k}\frac{1}{|D(a(s))|+b_t}.$$

Hence,

$$\sum_{t=1}^{\binom{N-2}{k-1}} \frac{1}{|D(a(s'))|+b_t} \leq \frac{M+k-1}{k} \sum_{t=1}^{\binom{N-2}{k-1}} \frac{1}{|D(a(s))|+b_t}$$

and finally,

$$\left(\sum_{t=1}^{\binom{N-2}{k-2}} \frac{1}{|D(a(s))|+a_t+|D(a(s'))|} + \sum_{t=1}^{\binom{N-2}{k-1}} \frac{1}{|D(a(s'))|+b_t}\right) \leq$$

$$\frac{M+k-1}{k}\left(\sum_{t=1}^{\binom{N-2}{k-2}} \frac{1}{|D(a(s))|+a_t+|D(a(s'))|} + \sum_{t=1}^{\binom{N-2}{k-1}} \frac{1}{|D(a(s))|+b_t}\right)$$

that gives

$$\frac{P(s')}{P(s)} \leq \frac{M+k-1}{k}.$$

□

**Lemma 1.** *Given $d \geq c$ and a finite sequence $\{b_j\}_1^T$ such that $0 < k-1 \leq b_j$ for all $j$, we have*

$$\frac{\sum_j \frac{1}{c+b_j}}{\sum_j \frac{1}{d+b_j}} \leq \frac{d+k-1}{c+k-1}.$$

*Proof.* Notice that for any $j$ we have $\frac{\frac{1}{c+b_j}}{\frac{1}{d+b_j}} \leq \frac{1/(c+k-1)}{1/(d+k-1)}$ because when $c \leq d$ it is a monotonically decreasing function of $b_j$. The desired result follows by summing up both sides over $j$ and dividing. □

Equation (1) shows that the sampling becomes more uniform as $k \to \infty$, and allows us to bound the uniformity of the output of Algorithm 2 as a function of $k$ (a pseudosolution cannot be much more likely than another one). For instance, when $M = 2$ and $k = 100$, a pseudosolution cannot be more than 1% more likely to be sampled than another one. Notice that larger values of $k$ would both improve the uniformity of the sampling and at the same time increase the number of output samples (sampling is without replacement). However, larger values of $k$ would also require more calls to the SAT solver.

**Remark**: When Algorithm 2 is used recursively as in Algorithm 1, the samples received as input are usually not truly uniform unless $k$ is larger than the total number of solutions[2]. Even though in general the sets $\Phi_i$ in Algorithm 1 do not meet the assumptions of Theorem 1, they tend to satisfy them as $k \to \infty$. The effect of rather small, practical values of $k$ is investigated empirically below, where we evaluate the statistical properties of the output samples $\Phi_L$.

### 4.3 Complexity

If $n$ is the number of variables of the input formula $F$, Algorithm 1 requires $O(\lceil \frac{n}{\log M}\rceil Mk)$ calls to the SAT solver to get at least $\min\{\#\text{solutions of F}, k\}$ samples. Larger values $M$ require more calls to the SAT solver, but intuitively also improve the uniformity of the sampling by reducing the number of recursions. In particular, in the extreme (impractical) case $M = 2^n$ we obtain truly uniform sampling because it corresponds to exact model counting (explicitly enumerating all solutions). Notice also that although in our experiments we use constant values for $k$ and $M$, they could be chosen as a function of the level $i$.

## 5 Evaluating Sampling Methods

It can be verified that Simulated Annealing (SA) is ergodic for all "temperatures" $T > 0$ when new configurations are generated by randomly flipping a variable chosen uniformly at random (so that the chains are irreducible [2]). Since according to the steady state probability distribution all satisfying assignments have the same probability, in principle SA will provide uniform samples if we are willing to wait until the stationary distribution is reached. However, the amount of time we have to wait until the chain reaches its steady state and between two consecutive samples is of paramount importance for any practical application. On the other hand, SampleSAT is often much faster at finding solutions because of the focused random walk component, but it does not provide any guarantee on the uniformity of the sampling. To evaluate the practical utility of these methods it is therefore important to quantitatively measure the uniformity of the samples provided after finite amounts of time.

The experiments are run on a set of formulas $F$ for which we know (analytically or using exact model counters [17]) that the number of distinct solutions $|S_F|$ is relatively small ($\leq 1000$). Therefore, by taking a sufficiently large number of samples, we are able to check the uniformity of the methods. In particular, let $N_i$ be the number of times the $i$th solution has been sampled, for $i = 1, \ldots, |S_F|$. Given $P$ samples from a truly uniform solution sampler, we expect $N_i$ to be close to $P/|S_F|$ (since for the law of large numbers $N_i/P$ converges to $1/|S_F|$ in probability as $P \to \infty$). To quantitatively measure the uniformity of the sampler, we use the Pearson's $\chi^2$ statistic defined as

$$\chi^2 = \frac{1}{P/|S_F|} \sum_i (N_i - P/|S_F|)^2,$$

where $P/|S_F|$ is the expected theoretical frequency under the hypothesis that the distribution is truly uniform. The $\chi^2$ value can be used to test a null hypothesis stating that the frequencies $N_i$ observed are consistent with a uniform distribution. In particular, the null hypothesis is rejected if the $\chi^2$ value is larger than a cutoff value that depends on the number of distinct solutions $|S_F|$ (specifying the number of degrees of freedom of the distribution) and on the statistical significance desired (e.g., 0.05).

---

[2] Uniformity is also guaranteed for the first $b$ levels, until $|S_b| < k$.

For our experiments, we use MiniSAT 2.2 [18] as a complete solver in Algorithm 1. Pseudosolutions are defined according to lexicographical variable ordering $\pi_{LX}$. For efficiency, all the experiments are run with $M = 2$. When we wish to obtain $P$ samples, we keep running SearchTreeSampler with a parameter $k < |S_F| < P$ until we obtain at least $P$ samples (taking exactly $k$ samples without replacement from $S_F$ per run), and we report the total running time. Note that choosing $k \geq |S_F|$ (e.g., $k = P$) wouldn't let us evaluate the performance of the method as an approximately uniform sampler, because it would correspond to enumerating all solutions (hence perfectly uniform sampling).

SA is evaluated as follows. We run the chain from a random initial truth assignment, discarding all the samples for the first $10^7$ steps (burn-in phase). After the burn-in phase, we assume that the Markov Chain has reached its steady state distribution and we start taking samples. Every $K$ steps, the current truth assignment $\sigma_t$ is taken as a sample from the steady state distribution. We wait for $K$ steps to ensure that consecutive samples are sufficiently independent. If the sample $\sigma_t$ is a solution ($\sigma_t \in S_F$), we output it, otherwise we discard it. This process is carried out until we find a prescribed number of solutions $P$ (non necessarily all distinct). SampleSAT is executed with default parameters, and all methods are run on the same 3GHz machine with 4Gb of memory.

# 6 Challenging Sample Spaces

## 6.1 Golf-Course energy landscape

We first consider a class of instances from [8] which we call $plateau(b)$, defined as

$$(x_1 \lor y_1) \land (x_1 \lor y_2) \land \cdots \land (x_1 \lor y_b) \land$$
$$(\neg x_1 \lor z_1) \land (\neg x_1 \lor \neg z_1).$$

This class of instances is hard for local search methods such as SA because it has a very large plateau of truth assignments (of size at least $2^{b+1}$) with energy $E(\sigma) = 1$, corresponding to assignments with $x_1 = 1$. The effect of the last two clauses is to enforce $x_1 = 0$, so that there are only two distinct solutions. Intuitively, this instance is difficult for SA because in order to reach one of the solutions, SA needs to set $x_1 = 0$. Setting $x_1 = 0$ is likely to violate several of the first $b$ clauses, making the uphill move unlikely to be accepted. Formally, in [8] it is shown (Theorem 1) that SA at fixed temperature with probability going to 1 takes time exponential (in $b$) to find a satisfying assignment.

As shown in [8], SampleSAT is not affected by energy plateaus because it uses focused moves to quickly reach solutions. Since it is not based on local information, SearchTreeSampler is not affected by energy plateaus. In fact, since $plateau(b)$ instances belong to 2-CNF (formulas with at most 2 variables per clause) and has two solutions, we can use the following result:

**Theorem 2.** *Algorithm 1 provides uniform samples in polynomial time for 2-CNF instances $F$ such that $S_F \leq k$.*

*Proof.* Follows from the fact that 2-SAT is solvable in polynomial time. Uniformity follows from the fact that Algorithm 1 will output all solutions when $S_F \leq k$. □

## 6.2 Energy barriers

We define an energy barrier between two variable assignments $\sigma, \sigma' \in \{0, 1\}^n$ as the minimum over the set of all possible paths on the Boolean hypercube $\{0, 1\}^n$ between $\sigma$ and $\sigma'$ of the maximum energy of the configurations in each path. Simulated Annealing is designed to deal with energy barriers (such as the ones encountered around a local minimum) by allowing occasional uphill moves (stochastic hill climbing), i.e. moves that lead to an increase of the energy function. In particular, larger values of the temperature parameter $T$ make the acceptance of uphill moves more likely.

In this section, we will show that SA can only deal with relatively low energy barriers if we want to maintain a reasonable efficiency in the rejection sampling scheme. Consider the following CNF formula that we call $XORBarrier(b)$:

$$(x_1 \Rightarrow y_1) \land (x_1 \Rightarrow y_2) \land \cdots \land (x_1 \Rightarrow y_b) \land$$
$$(\neg x_1 \Rightarrow \neg y_1) \land (\neg x_1 \Rightarrow \neg y_2) \land \cdots \land (\neg x_1 \Rightarrow \neg y_b).$$

We use the term $XORBarrier(b)$ because it is equivalent to a conjunction of XOR constraints of the form $\neg(x_1 \; XOR \; y_1) \land \cdots \land \neg(x_1 \; XOR \; y_b)$. It is easy to see that for any value of $b$, these instances have only two solutions $(x_1, y_1, \cdots, y_b) = (0 \ldots 0)$ and $(x_1, y_1, \cdots, y_b) = (1 \cdots 1)$. It can be seen that $d(i) = |\{\sigma \in \{0, 1\}^n | E(\sigma) = i\}| = 2\binom{b}{i}$. For any value of the temperature parameter $T$, the steady state probability of having SA in a configuration with energy $i$ is

$$\mathbb{P}[E(\sigma) = i] = \sum_{\sigma | E(\sigma) = i} P_T(\sigma) \propto n(i) e^{-\frac{i}{T}} = 2\binom{b}{i} e^{-\frac{i}{T}}.$$

The key feature of this type of instances is the following. Since Simulated Annealing (and Gibbs Sampling) flips only one variable at a time, it is easy to see that any path in the Markov Chain graph that brings from one solution to the other one will contain a configuration that violates $b/2$ constraints. Therefore we conclude that there is an energy barrier of "height" $b/2$ between the two solutions.

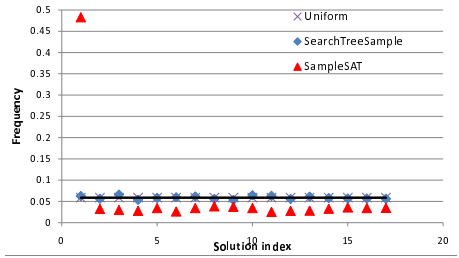 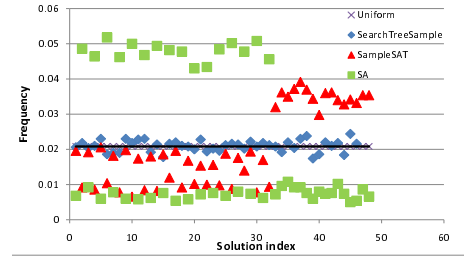

(a) Asymmetric space with energy barrier.  (b) Random formula with energy barrier.

Figure 2: Uniformity of sampling. Frequency each solution is sampled by different methods.

For large values of $b$, one cannot find a temperature $T$ that provides both high enough $P_T(0)$ (i.e. small number of flips per solution) and $P_T(b/2)$ (i.e. probability of "jumping" over the barrier) to obtain a practical sampler. For instance, for $b = 80$, to get $P_T(40) > 10^{-9}$ (i.e. climbing the barrier on average once in about a billion samples) SA needs a temperature $T > 0.75$, but $P_{0.75}(0) = 7.4 \times 10^{-9}$ (which means that on average we need more than $10^8$ samples to get a single solution). Similar results are obtained when using Gibbs sampling. This is because a Gibbs sampler has the same Boltzmann steady state probability distribution and it also proceeds by flipping a single variable at a time, so it cannot "jump" over the barrier with a single move.

Since instances in $XORBarrier(b)$ belong to 2-SAT and have 2 solutions, from Theorem 2 Algorithm 1 provides uniform samples in polynomial time.

A common strategy to partially overcome the ergodicity problems of Gibbs sampling and Simulated Annealing is the use of multiple parallel chains [9] or restarts [8]. These approaches can be quite effective and, for instance, are capable of producing uniform samples for the $XORBarrier(b)$ class of instances presented above. Intuitively, this is because on average $50\%$ of the random initial assignments will be on one side of the barrier (i.e., the half-hypercube with $x_1 = 0$) and the other $50\%$ on the other side (i.e., $x_1 = 1$).

### 6.3 Asymmetric spaces with energy barriers

These approaches are however not sufficient to sample from distributions where there are "solution clusters" (i.e., groups of solutions that are close in Hamming distance) of different sizes that are separated by energy barriers. Consider for instance the following class of instances that we call $AsymXORbarrier(b, \ell)$:

$$XORBarrier(b) \wedge (x_1 \vee z_1) \wedge (x_1 \vee z_2) \wedge \cdots \wedge (x_1 \vee z_\ell).$$

The effect of the additional clauses $(x_1 \vee z_i)$ is to create a cluster of $2^\ell$ solutions on one side of the barrier ($x_1 = 1$), while there exist only one solution with $x_1 = 0$.

As shown in Figure 2a and Table 1 (bottom rows), SearchTreeSampler provides a much more uniform sampling compared to SampleSAT on this type of instances. Specifically, the uniformity of sampling hypothesis is rejected for SampleSAT according to the $\chi^2$-test. This is because SampleSAT employs restarts and therefore samples the first solution in Figure 2a (corresponding to the isolated solution with all variables set to 0) too often, i.e. about $50\%$ of the times. Similarly biased results are obtained using a Gibbs Sampler with multiple parallel chains (as implemented in the Alchemy system [9]), while plain SA or Gibbs is not able to cross the barrier in a reasonable amount of time. Further, we see from Table 1 (4th column) that thanks to MiniSAT's reasoning power, SearchTreeSampler is about 3 orders of magnitude faster than SampleSAT.

### 6.4 Embedded energy barriers

In Table 1 (top rows), we evaluate the performance of the methods considered on three types of instances: a logistic one generated by SATPlan [19] (*logistic*, with 110 variables, 461 clauses, and 512 solutions), a graph Coloring problem from SatLib [20](*coloring*, with 90 variables, 300 clauses, and 900 solutions), and a random 3-SAT formula (*random*, with 75 variables, 315 clauses, and 48 solutions). We collect respectively $P = 50000$, $P = 200000$, and $P = 5000$ samples for each instance. We also artificially introduce an energy barrier in each of these instances by choosing a variable $z$ such that there are roughly half solutions with $z = 1$ and half solutions with $z = 0$ in the original formula. We then introduce a new set clauses defined by $XORbarrier(40)$ where we substitute $z$ to $x_1$ and $y_1, \ldots, y_{40}$ are fresh variables. Note that this does not change the number of solutions.

We experimented with several values of $k$ and $T$ and we provide a summary of the best results obtained in Table 1, both for the original formulas and the ones with an embedded energy barrier. By carefully choosing the tempera-ture parameter $T$, SA can provide uniform samples for the original formulas. However, it is very slow compared to the other methods (smaller $T$ would make it faster, but the samples would not be uniform). On the other hand, SampleSAT is generally much faster, but the samples provided are far less uniform. SearchTreeSampler is superior to both SA and SampleSAT, both in terms of running time (at least 2 orders of magnitude faster) and uniformity. As expected, increasing $k$ improves the uniformity and it does not affect the running time because it also produces more samples per run.

| Method | Instance | Parameter | Time (s) | $\chi^2$ | P-value |
|---|---|---|---|---|---|
| SA | logistic | T=0.25 | 11028 | 550 | **0.11** |
| SampleSAT | logistic | - | 7598 | 534.9 | **0.23** |
| SearchTreeSampler | logistic | k=5 | 9.8 | 545.82 | **0.14** |
| SearchTreeSampler | logistic | k=20 | 10.1 | 487.43 | **0.77** |
| SearchTreeSampler | logistic | k=50 | 9.6 | 407.01 | **0.99** |
| SA | logistic+Barrier | T=0.25 | 42845 | 50495.8 | 0 |
| SampleSAT | logistic+Barrier | - | 7296 | 178860.2 | 0 |
| SearchTreeSampler | logistic+Barrier | k=20 | 42.53 | 469.23 | **0.90** |
| SA | coloring | T=0.25 | 7589 | 875.131 | **0.71** |
| SampleSAT | coloring | - | 28998 | 132559 | 0 |
| SearchTreeSampler | coloring | k=50 | 184 | 844 | **0.90** |
| SearchTreeSampler | coloring | k=100 | 204 | 808.3 | **0.98** |
| SearchTreeSampler | coloring | k=200 | 228 | 672.15 | **1.00** |
| SA | coloring+Barrier | T=0.25 | 29434 | 100905 | 0 |
| SampleSAT | coloring+Barrier | - | 29062 | 141027 | 0 |
| SearchTreeSampler | coloring+Barrier | k=100 | 435 | 746.40 | **0.99** |
| SA | random | T=0.3 | 8673 | 36.62 | **0.88** |
| SampleSAT | random | - | 740 | 970.21 | 0 |
| SearchTreeSampler | random | k=10 | 2.5 | 76.84 | 0 |
| SearchTreeSampler | random | k=15 | 3 | 31.28 | **0.96** |
| SearchTreeSampler | random | k=20 | 5 | 29.26 | **0.98** |
| SA | random+Barrier | T=0.3 | 71077 | 4211.40 | 0 |
| SampleSAT | random+Barrier | - | 744 | 1104.9 | 0 |
| SearchTreeSampler | random+Barrier | k=10 | 7.0 | 64.55 | 0.04 |
| SearchTreeSampler | random+Barrier | k=15 | 7.1 | 32.64 | **0.94** |
| SearchTreeSampler | random+Barrier | k=20 | 7.1 | 32.33 | **0.95** |
| SampleSAT | AsymXORBarrier(80,4) | - | 290 | 6508 | 0 |
| SearchTreeSampler | AsymXORBarrier(80,4) | k=2 | 0.7 | 51.54 | 0 |
| SearchTreeSampler | AsymXORBarrier(80,4) | k=5 | 0.4 | 17.35 | **0.36** |
| SearchTreeSampler | AsymXORBarrier(80,4) | k=10 | 0.3 | 7.84 | **0.95** |
| SampleSAT | AsymXORBarrier(80,8) | - | 4260 | 1893391 | 0 |
| SearchTreeSampler | AsymXORBarrier(80,8) | k=25 | 1.9 | 220.14 | **0.94** |
| SearchTreeSampler | AsymXORBarrier(80,8) | k=50 | 1.6 | 197.17 | **0.99** |

Table 1: Uniformity of sampling for instances with and without embedded energy barriers. P-value is the probability of observing an event at least as extreme under the null hypothesis (uniform sampling). In bold null hypothesis is not rejected.

As expected, the introduction of energy barriers severely limits the ergodicity of SA, as shown in the last column of Table 1 and in Figure 2. As a result, both SA and SampleSAT fail to provide uniform samples (according to the $\chi^2$ test) for the instances with embedded energy barriers. On the other hand, the uniformity of SearchTreeSampler is not affected, although the runtime increases (because the search tree is deeper with the introduced fresh variables).

## 7 A New Model Counting Technique

A typical way [21, 22] to estimate the number of solutions $|S_F| = Z$ (model count) of a formula $F$ using a sample approximation is to estimate a series of multipliers

$$\frac{Z}{Z(x_1 = a_1)} \frac{Z(x_1 = a_1)}{Z(x_1 = a, x_2 = a_2)} \cdots \frac{Z(x_1 = a_1, \cdots, x_{n-1} = a_{n-1})}{1}.$$

For instance, $\frac{Z(x_1=a_1)}{Z}$ is given by the fraction of solutions that have $x_1 = a_1$ (the number of solutions in the green vs. yellow part of the search tree in Figure 3), and so on. This can be a challenging task because one has to heuristically select a "good" variable ordering and polarities $a_1, \cdots, a_n$ to condition on, and solve multiple sampling tasks with some of the variables clamped [22].

Algorithm 1 provides a new direct way of estimating the number of solutions. Let $S_i$ be the set of pseudosolutions of level $i$. Suppose $Z = |S_n| > 0$ (the formula is satisfiable), and for simplicity let $M = 2^\ell = 2$. Then, we have

$$Z = |S_n| = \frac{|S_n|}{|S_{n-1}|} \frac{|S_{n-1}|}{|S_{n-2}|} \frac{|S_{n-2}|}{|S_{n-3}|} \cdots \frac{|S_1|}{1}$$

(see right panel of Figure 3), where we can estimate each multiplier using the relation

$$\frac{|S_i|}{|S_{i-1}|} = \frac{1}{|S_{i-1}|} \sum_{s \in S_{i-1}} |D(s)| \approx \frac{1}{|\Phi_{i-1}|} \sum_{\tilde{s} \in \Phi_{i-1}} |D(\tilde{s})|,$$

where elements $\tilde{s} \in \Phi_{i-1}$ provided by Algorithm 1 are (approximately) uniformly sampled elements of $S_{i-1}$. Note that each multiplier corresponds to the average number of descendants at a given level $i$ of the search tree, and we can estimate all of them in a single run of SearchTreeSampler (i.e. no need to solve multiple sampling tasks as in [22]).

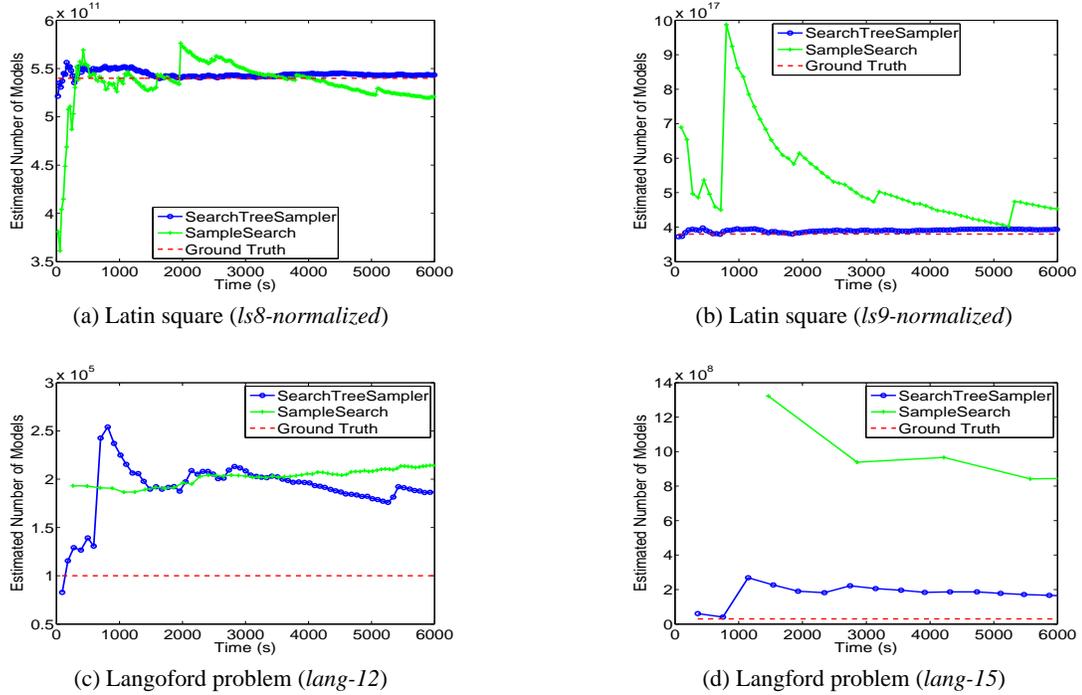

Figure 4: Estimated model count as a function of runtime.

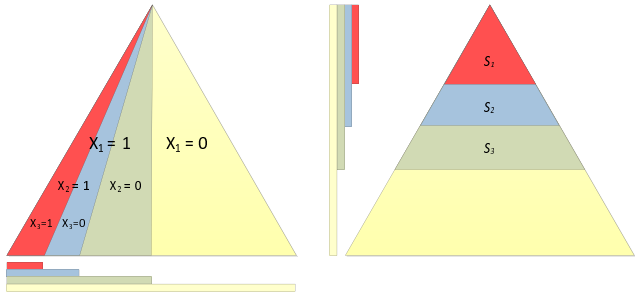

Figure 3: Pictorial representation of the traditional (left) and new model counting techniques (right). Notice the vertical vs. horizontal division of the search tree.

### 7.1 Experimental Results

We compare our new counting method with SampleSearch [23], currently the best model counter based on (importance) sampling [12], on several large instances (with known ground truth) from a standard model counting benchmark [3]. In Figure 4 we plot the estimated model count over time (as more samples are being collected) for the two methods, for 4 representative instances (encodings of Latin Squares and Langford's problems). Note that both methods rely on MiniSAT as internal constraint solver. Both methods are run on the same 3GHz machine, SearchTreeSampler with $k = 500$ and SampleSearch with parameters as in [23].

Our empirical results show that SearchTreeSampler converges faster and provides more accurate estimates. Similar results are obtained for other instances in the benchmark (not reported for space reasons). Note that these instances have a rather large number of solutions so it is difficult to evaluate the uniformity of the sampling using a $\chi^2$ test. In this case, the accuracy of the estimates of $Z$ is an indication of the of the quality of the sampling scheme.

## 8 Conclusions

We presented SearchTreeSampler, a new method to sample solutions of Boolean formulas and to estimate their count. Our method uses a constraint solver as a blackbox, and can therefore leverage the reasoning power of state-of-the-art constraint solving technology, while maintaining an approximately uniform exploration of the search tree. We presented several challenging domains, such as energy barriers and asymmetric search spaces, that reveal the difficulties of sampling. We presented data demonstrating that standard methods such as Simulated Annealing, Gibbs Sampling, and SampleSAT seriously suffer from these difficulties in terms of runtime and uniformity of the sampling. Our results also demonstrate that SearchTreeSampler can overcome many of these difficulties and is considerably faster on a set of benchmark problems. Further, we show that the intermediate samples of pseudosolutions provided by our method can be used in a new model counting technique, which is shown to be very effective in practice.